\documentclass{article}
\usepackage{microtype}
\usepackage{graphicx}
\usepackage{subfigure}
\usepackage{booktabs} 
\usepackage{multirow}

\newcommand{\TODO}[1]{\textcolor{black}{#1}}

\usepackage{hyperref}

\usepackage[accepted]{icml2024}

\usepackage{amsmath}
\usepackage{nccmath}
\usepackage{amssymb}
\usepackage{mathtools}
\usepackage{amsthm}
\usepackage{stfloats}
\usepackage[capitalize,noabbrev]{cleveref}

\theoremstyle{plain}

\theoremstyle{definition}

\theoremstyle{remark}

\usepackage[textsize=tiny]{todonotes}
\usepackage{caption}

\icmltitlerunning{Prompt-based Visual Alignment for Zero-shot Policy Transfer}

\begin{document}

\twocolumn[
\icmltitle{Prompt-based Visual Alignment for Zero-shot Policy Transfer}




\begin{icmlauthorlist}
\icmlauthor{Haihan Gao}{ustc,skl}
\icmlauthor{ Rui Zhang}{skl}
\icmlauthor{Qi Yi}{ustc}
\icmlauthor{Hantao Yao}{ia}
\icmlauthor{Haochen Li}{is,ucas}
\icmlauthor{Jiaming Guo}{skl}
\icmlauthor{Shaohui Peng}{is}
\icmlauthor{Yunkai Gao}{ustc,skl}
\icmlauthor{QiCheng Wang }{skl,ucas}
\icmlauthor{Xing Hu}{skl,sh}
\icmlauthor{Yuanbo Wen}{skl}
\icmlauthor{Zihao Zhang}{skl}
\icmlauthor{Zidong Du}{skl,sh}
\icmlauthor{Ling Li}{is,ucas}
\icmlauthor{Qi Guo}{skl}
\icmlauthor{Yunji Chen}{skl,ucas}
\end{icmlauthorlist}
\icmlaffiliation{skl}{SKL of Processors, Institute of Computing Technology, CAS}
\icmlaffiliation{is}{Institute of Software, Chinese Academy of Sciences}
\icmlaffiliation{sh}{Shanghai Innovation Center for Processor Technologies}
\icmlaffiliation{ustc}{University of Science and Technology of China}
\icmlaffiliation{ucas}{University of Chinese Academy of Sciences, China}
\icmlaffiliation{ia}{Institute of Automation, Chinese Academy of Sciences}

\icmlcorrespondingauthor{Yunji Chen}{cyj@ict.ac.cn}
\icmlkeywords{Machine Learning, ICML}

\vskip 0.3in
]


\printAffiliationsAndNotice{}  
\newcommand\haihan[1]{\textcolor{black}{ #1}}
\newcommand\remove[1]{\textcolor{blue}{(remove: #1)}}
\newcommand\yq[1]{\textcolor{red}{#1}}
\begin{abstract}
Overfitting in RL has become one of the main obstacles to applications in reinforcement learning(RL).
Existing methods do not provide explicit semantic constrain for the feature extractor, hindering the agent from learning a unified cross-domain representation and resulting in performance degradation on unseen domains. Besides, abundant data from multiple domains are needed.
To address these issues, in this work, we propose prompt-based visual alignment (PVA), a robust framework to mitigate the detrimental domain bias in the image for zero-shot policy transfer. 
Inspired that Visual-Language Model (VLM) can serve as a bridge to connect both text space and image space, we leverage the semantic information contained in a text sequence as an explicit constraint to train a visual aligner. Thus, the visual aligner can map images from multiple domains to a unified domain and achieve good generalization performance. To better depict semantic information, prompt tuning is applied to learn a sequence of learnable tokens.
With explicit constraints of semantic information, PVA can learn unified cross-domain representation under limited access to cross-domain data and achieves great zero-shot generalization ability in unseen domains.
We verify PVA on a vision-based autonomous driving task with CARLA simulator. Experiments show that the agent generalizes well on unseen domains under limited access to multi-domain data.

\end{abstract}    
\section{Introduction}

\begin{figure}[htbp]
\centering
\includegraphics[width=\columnwidth]{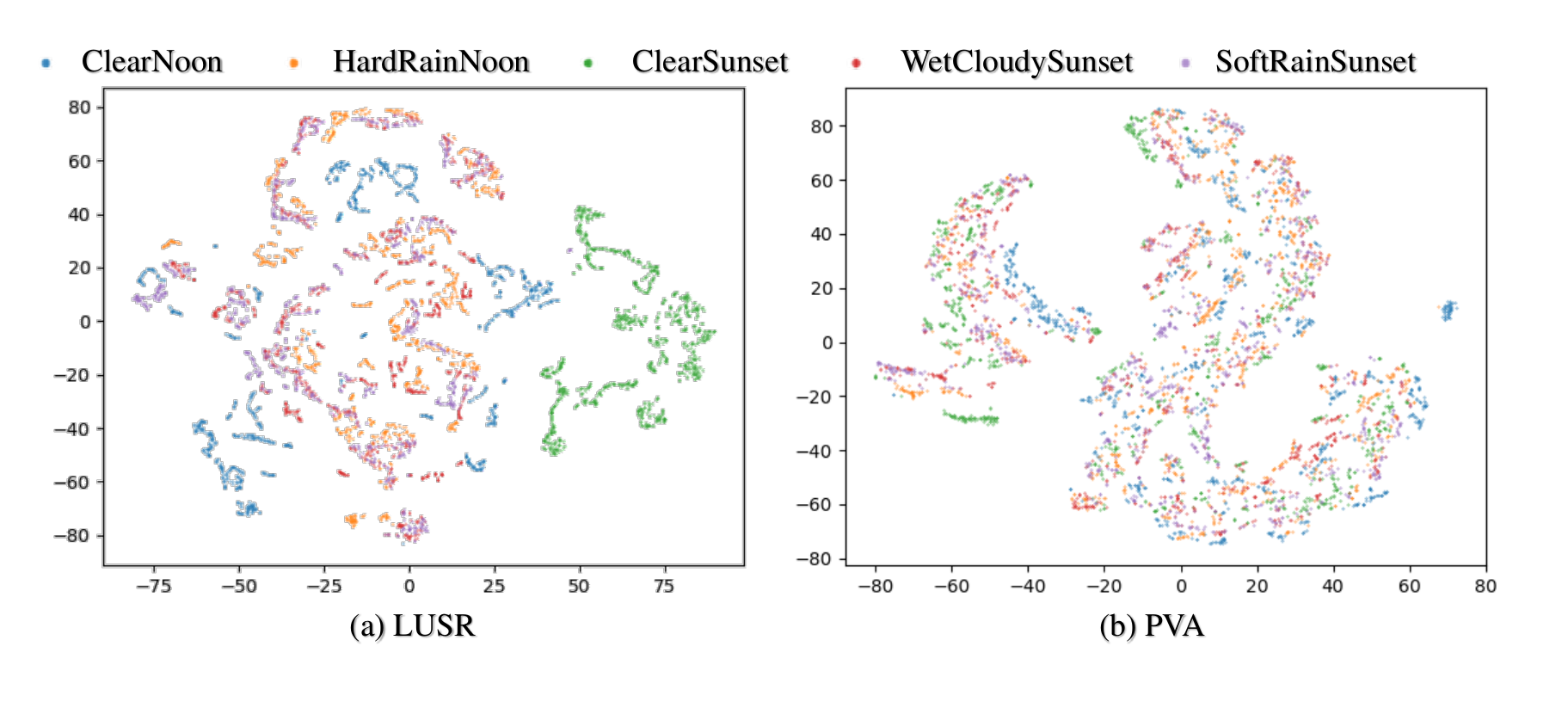}
\vspace{-1em}
\caption{Input embedding distributions for the policy before (a) and after (b) domain alignment. Different colors represent different domains, where ClearNoon and HardRainNoon are applied in the training set. (a) Latent features generated by LUSR demonstrate severe out-of-distribution phenomena between seen and unseen domains (such as ClearSunset). (b) our approach mitigates the domain bias across various domains and well aligns the latent distributions between training and testing domains.}

\label{fig:visualemb}
\end{figure}

Reinforcement Learning(RL) has enabled remarkable success in many control tasks, such as robot manipulation, video games, and autonomous driving. However, overfitting in the training environment has become a prominent problem, where RL agents can not tolerate slight shifts between train and test domains. Aside from this, interaction with test  environments can be costly. Both challenges make it essential to investigate a strategy to zero-shot transfer a policy without access to data from test domains.



Many methods have been proposed to transfer a policy under shifts in observations. LUSR\cite{LUSR} and DARLA\cite{darla} train a transferable policy via learning a generalizable representation from different states, where variation-auto-encoder\cite{Variatio-Auto-Encoder} is applied to map states from different domains into aligned representations. Image-to-image translation model serves as a map from one domain to another domain\cite{transferimagerl,virtualimage}. Instead of transforming the original observation space to another unified space, domain randomization\cite{Domaingeneralization} and data augmentation technique\cite{DrQ,CURL} are applied to improve the generalization ability of RL agents by promoting data diversity via image augmentations during the training stage. 

Although these works have demonstrated compelling adaptation performance over different domains, 
there are still two problems existing in these methods, namely, the lack of explicit semantic constraints and the requirements of sufficient multi-domain data, which hinder the generalization in complex scenarios. 
During extracting aligned features among different training domains with a feature extractor, these methods do not constrain the feature extractor explicitly on semantic information, which hinders the agent from learning a unified cross-domain representation. 
The representation learned by these works might contain domain bias of the training domains and will cause performance degradation on unseen domains. 
As shown in Figure \ref{fig:visualemb}(a), the representation distribution of the testing domains does not match that of the training domains. 
Besides, all of these methods require abundant data from multiple domains to adapt to a new domain, which can be expensive and even unavailable sometimes.


To alleviate these issues, we propose a framework called Prompt-based Visual Alignment (PVA) for zero-shot policy transfer.
The key point of PVA is to leverage the semantic information contained in a text sequence as an explicit constraint to train a visual aligner, aiming to obtain high generalization and remove domain bias.
The visual aligner can achieve cross-domain alignment by mapping the images of different domains to a unified domain.
Inspired by the remark generalization ability of the pretrained Visual-Language Model(VLM), we use VLM to bridge textual and visual modalities.
Considering the semantic information is contained in a sequence of text, we can use VLM as the explicit constraint of retaining semantic information during the alignment of mapping images.
To better depict semantic information, prompt tuning can be applied to learn a sequence of learnable tokens instead of the fixed template.


\begin{figure*}[t]
    \centering
    \includegraphics[width=0.85\linewidth]{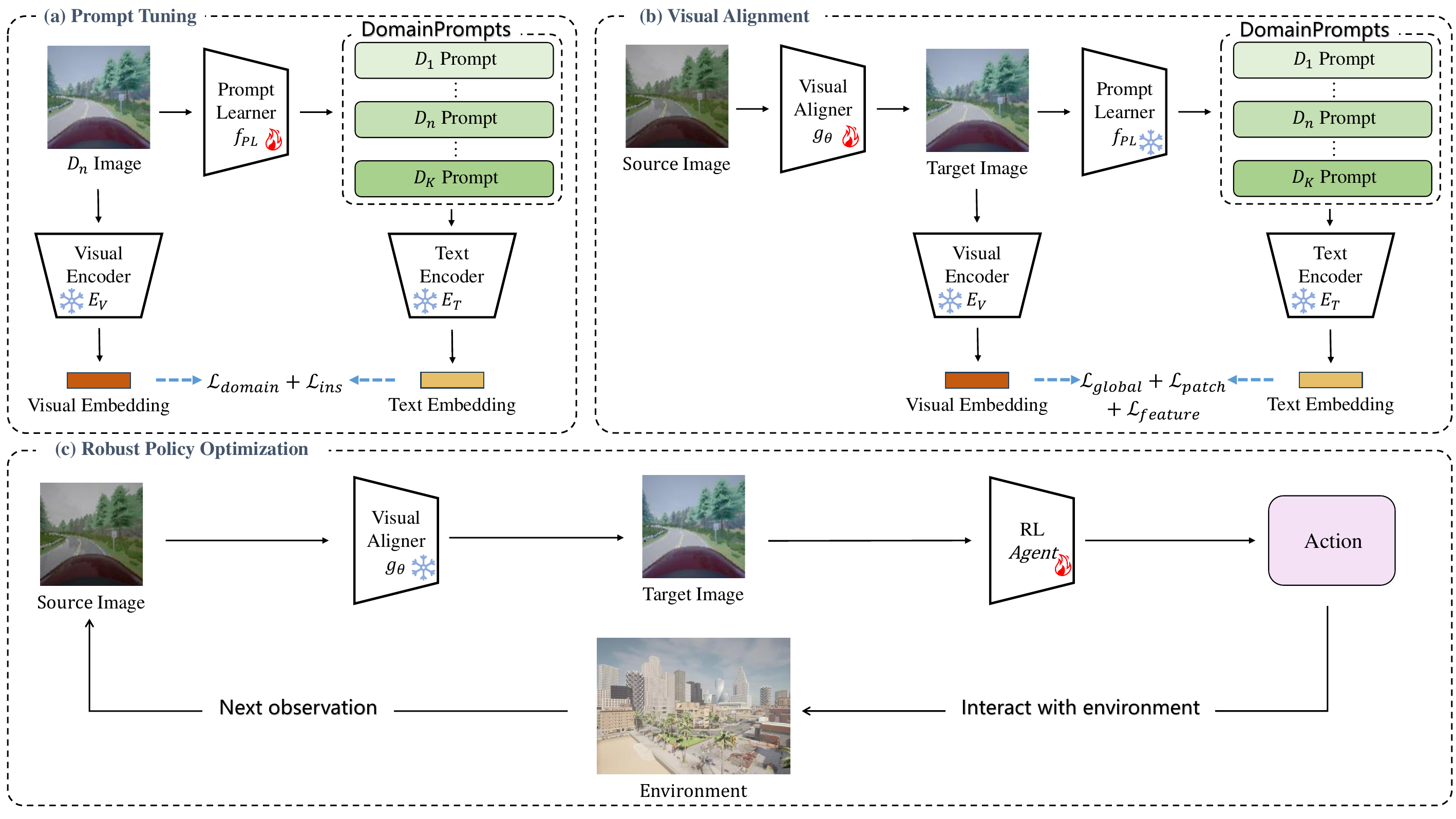}

  \caption{ {\textbf{Overview of Prompt-based Visual Aligner(PVA)}.There are two key components in our methods. Prompt Learner $f_{PL}$ to obtain a learnable prompt from the input image. The visual aligner $g_\theta$ will transfer the image from one domain to another domain via the semantic information contained in the learnable prompt. Then the agent applies the transferred image to train a robust policy. In the illustration, we use \includegraphics[width=0.02\textwidth]{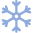} to indicate the network is frozen and \includegraphics[width=0.02\textwidth]{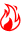} represents the network will be updated during training.}}  
  \label{fig:pipeline}
\end{figure*}

Based on the above analysis, the proposed PVA consists of three stages: Prompt Tuning, Visual Alignment, and Robust Policy Optimization. 
In the Prompt Tuning stage, we leverage the sequence of learnable tokens to obtain an accurate description of semantic information in the input image by aligning the text embedding with the visual embedding.
In Visual Alignment stage, a visual aligner is trained to map the images of multiple training domains to a unified domain. We provide an explicit constraint of retaining the semantic information by aligning the mapping image with prompts generated in the first phase. 
In Robust Policy Optimization stage, we apply the visual aligner from the second stage to train a robust RL agent. 
As shown in Figure \ref{fig:visualemb}(b), with explicit constraints of relating semantic information from VLM, PVA can learn unified cross-domain representation from various domains and achieves great zero-shot generalization ability in unseen domains.
Moreover, the visual aligner of PVA can be learned under limited amount of data from various domains.

We verify the proposed PVA with CARLA simulator\cite{CARLA}. Each domain refers to a specific type of weather, which differs in illumination conditions and precipitation. Experiments have shown that PVA has a good zero-shot transfer ability over unseen domains with limited data.

\section{Related Work}

\textbf{Domain Adaptation(DA) in RL:} Domain Adaptation in Reinforcement Learning aims to train an agent to generalize in different environments. 
Most current approaches focus on how to deal with the shift in observation space, especially in a visual-based task. 
Current methods improve the adaptation ability of RL agents from three aspects: Domain randomization with diverse data, representation learning to learn domain-agnostic representation across different domains, and image-to-image translation.



Domain randomization involves multiple domains during training to improve the robustness of agents\cite{Domaingeneralization,sim2real,Sim2RealDynamicrand}. 
However, since access to multiple domains is not always available, data augmentation approaches are proposed to increase domain diversity manually.
{Data-augmentation-based methods, including RAD\cite{rlwithaugdata} and DrQ\cite{DrQ}, are proposed to increase data diversity based on image transformations.} RAD indicates that simple augmentation methods such as random cropping or gray scaling will improve the generalization ability of agents.{DrQ provides a plug-and-play data augmentation method for various RL algorithms.}
\haihan{Arbrarlity selected augmented method might break semantic information in the observation, which will mislead the agent to do the appropriate action.} 

Aside from domain generalization, some researchers focus on utilizing representation learning techniques to promote the robustness of RL agents, such as CURL\cite{CURL}, LUSR\cite{LUSR}, OPA\cite{OPA} and DARLA\cite{darla}. {CURL aims to learn invariant representation via augmented data. DARLA and LUSR apply $\beta$-VAE\cite{betavae} to learn domain-agnostic representations from several domains.} However, these methods require plenty of data from various domains as well as neglecting the semantic information of different parts in the observations, which will hinder the encoder from extracting domain-agnostic representation in complex scenarios.

Some other approaches try to construct the map between two domains by image-to-image translation\cite{cyclegan,transferimagerl,virtualimage,styleagnostic}. These approaches utilize GAN\cite{GAN} to generate the observation from one domain to another domain. However, there exists instability in the training of GAN. It also requires data from the source domain, which violates the zero-shot condition.

Our approach builds a generalizable image-to-image mapping from a small set of data collected in different domains. Traditional image-to-image methods fail in our setting because of overfitting in such a small dataset and fail to generalize on unseen domains. 


\textbf{Visual Language Model(VLM):} VLM demonstrates great multimodal data encoding capabilities, as its encoding result has demonstrated exceptional performance across various downstream tasks. CLIP\cite{CLIP} and BLIP\cite{li2022blip} align visual embedding with text embedding. Two achievements in deep learning contribute to VLM greatly: attention-based transformer and unsupervised contrastive learning.
To obtain enough amount of data to pretrain the VLM, a web-scale dataset containing plenty of image-text pairs is applied. By aligning text and visual embeddings, VLM endows the visual embedding semantic information.

VLM consists of two parts, the text encoder $E_T$ and the visual encoder $E_V$, which map the texts and visual images to latent representation respectively. The better text describing the image, the higher cosine similarity between text and visual embedding will be obtained.

\textbf{Prompt Tuning:} Prompt tuning is proposed to improve the downstream task performance of VLM. It will be costly to fine-tune or train a VLM from scratch. Prompt tuning, such as CoOp\cite{CoOp}, CoCoOp\cite{CoCoOp}, and KgCoOp\cite{KgCoOp,DAPro,tcp}, exhibits obvious improvement on various visual downstream tasks without changing the parameters of VLM. 
CoOp is the first work to treat the prompt as an optimizable sequence of parameters. CoCoO adds the instance prompts to enhance the generalization ability of prompts over unseen classes. CoCoOp is proposed to improve the generalization ability of prompts via instance prompts.
Motivated by CoOp and CoCoOp, we leverage the global and domain-specific prompts to obtain a more accurate description of all images in the domain and their shared attribute. Instance prompt is added to generalize to unseen domains better as well.

\section{Preliminary}

Reinforcement Learning aims to learn to take actions to maximize the cumulative rewards. The environment is formulated as a Markov Decision Process (MDP) $(S,A,T,R)$ where $S$ is the state space, $A$ is the action space. $T$ is the dynamic transition function and $R$ is the reward function. 

To formulate the policy transfer scenarios in reinforcement learning, we define $K$ training domains $D_1,D_2,\cdots,D_K$. Each domain $D_i$ corresponds to an MDP $(S_i,A_i,T_i,R_i)$. Different domains share the same action space $A_1 = A_2=\cdots =A$ but have different observation spaces.

In our approach, we sampled two datasets from the MDPs. The first dataset $D_{semantic}$ is sampled over $K$ training domains with $M$ images per domain to train a map from multiple domains to one unified domain which will reduce the domain bias. 

The second dataset $D_{policy}$ is generated by sampling $N$ images from one of the $K$ training domains and is used for the downstream control task.
\haihan{$D_{semantic}$ samples mini-batch from multiple domains and the agent is trained solely with one domain, which will reduce the sampling cost.} 

\section{Prompt-based Visual Aligner}
In this work, we propose a Prompt-based Visual Aligner (PVA) framework to train a visual aligner for zero-shot policy transfer. The visual aligner serves as an image-to-image mapping from an arbitrary domain to a unified domain to reduce domain bias.
\haihan{PVA consists of three stages. 1) Prompt tuning: Train the prompt learner to obtain a fine-grained description of every image, which is illustrated in Figure \ref{fig:pipeline}(a). The prompt learner takes an image $I$ and outputs $K$ learnable prompts $\{P_I^k, k=1, 2, \cdots, K\}$.  2) Visual aligning: Train the visual aligner $g_\theta$ which is an image-to-image map with parameter $\theta$ to align different domains, illustrated in Figure~\ref{fig:pipeline}(b). $g_\theta$ is optimized by aligning the visual and prompt semantic information contained in the learnable prompts $\{P_I^k\}$. 
3)Robust policy training: Train the agent by applying transferred images converted by $g_\theta$ to obtain a robust policy against observation shifts illustrated in Figure \ref{fig:pipeline}(c). }

In the first two stages, only a relatively small dataset containing hundreds of images from several domains is engaged. We use CLIP as the bridge between text descriptions and image inputs. 

\label{sec:formatting}

\subsection{Prompt Tuning}
\haihan{Visual aligner $g_\theta$ is optimized by matching the visual embedding generated by VLM's visual encoder $E_V$ and the text embedding generated by VLM's text encoder $E_T$, which will be beneficial from a precise text description. We apply prompt tuning to generate a disentangled and precise description for each image. To fulfill this target, we separate the prompt into different parts to depict domain-shared, domain-specific, and instance-conditional level information respectively. In this part, we will detail the composition of prompts as well as how to learn each part.}


\subsubsection{Prompt Design}
Inspired by prompt tuning techniques \cite{CoCoOp,CoOp}, we utilize a sequence of learnable tokens to describe semantic information in the image. The design of prompts needs to satisfy several requirements to obtain accurate and generalizable descriptions.
1) The prompt has to contain the global semantic information shared across different domains, which depicts the attribute of the task.
2) The prompt has to contain domain-specific information to distinguish various domains. It describes the discrepancy between domains that we want to neglect during training the agent.
3) The prompt should be able to distinguish instances within the same domain, which describe essential information for the downstream control and should be kept by $g_\theta$.

Corresponding to these requirements, we design an effective domain-adaptive instance-conditional prompt, which contains different sections to fulfill these objects:
\begin{align}
    P_I^k&=[P_F^1][\textcolor[rgb]{1,0,0}{P_C}][\textcolor[rgb]{0,0,1}{P_S^k}][\textcolor[rgb]{0,1,0}{P_G}] [P_F^2],
    \label{eq:ensemble}
\end{align}
$k=1,2,\cdots,K$, where $P_I^k$ implies the $k$-th prompt generated by Image $I$, and $P_F^1,P_F^2$ are fixed templates such as "Driving the car on" and "the day.". 

Three learnable components are involved in $P_I^k$. 
The global prompt $\textcolor[rgb]{0,1,0}{P_G = [v_g^1][v_g^2]\cdots [v_g^{L_G}]}$ is shared across different domains, which represents knowledge of the whole task. 
The domain-specific prompt $\textcolor[rgb]{0,0,1}{P_S^k = [v_{s}^{k,1}][v_{s}^{k,2}]\cdots [v_{s}^{k,L_S}]}$ represents specific knowledge of each domain. $K$ domain-specific prompts correspond to $K$ training domains contained in $D_{semantic}$. 
To distinguish different images within the same domain, we use a neural network $h_{\phi}$ to learn instance-conditional prompt $\textcolor[rgb]{1,0,0}{P_C = h_\phi(I) = [v_C^1][v_C^2]\cdots [v_C^{L_C}] } $, where $\phi$ is the parameter of the neural network. $P_G\in R^{L_G\times d},P_S\in R^{K\times L_S\times d}$ are optimizable parameters while $P_C$ is extracted by the neural network $h_\phi$.
For $K$ prompts $P_I^1,P_I^2,\cdots,P_I^K$ generated by Image $I$, they share the same global prompts as well as the instance conditional prompts and differ in domain-specific prompts.
\subsubsection{Tuning the learnable parts of prompt}
The prompts are optimized separately: 1) Tune the domain-specific and global prompts. 2) Tune the instance-conditional prompts. Both parts are learned by contrastive loss.

{Tune the global and domain-specific prompts:}
Given image $I$ of domain $c$, we aim to match the corresponding prompt $P_I^c$ with image $I$ contrasting to prompt $P_I^i,i\neq c$ of other domains.
Specifically, we calculate the matching by cosine similarity between visual embedding from visual encoder $E_V$ and text embedding from textual encoder $E_T$, which will map images and prompts into the unified embedding space.

The domain-specific and global prompts are optimized by $L_{domain}$, which is formulated in Eq \eqref{eq:lossglobal} 
\begin{equation}\label{eq:lossglobal}
    \resizebox{0.9\hsize}{!}{%
        $L_{domain}(I,c) = -\log \frac{\exp{(\cos(E_V(I),E_T({P_I}^c))/\tau_d)}}{\sum_{i=1}^K \exp{(\cos{(E_V(I),E_T(P_I^i)}/\tau_d)}},$ %
        }
\end{equation}

where $\tau_{d}$ is the temperature parameter. 

{Tune the instance prompts:}
We tune the instance-conditional prompt by comparing the prompts generated by the images in the same domain. 
Given $B$ images $I_1,I_2,\cdots,I_B$ in the $c$-th domain from $D_{semantic}$, their corresponding prompts are denoted as  $P_{I_1}^c,P_{I_2}^c,\cdots,P_{I_B}^c$.
Matching the $I_k$ with corresponding $P_{I_k}^c$ will make the instance-conditional prompt distinguish between different images within the same domain.
The instance-conditional loss $L_{ins}$ is formulated in Eq \eqref{eq:lossinstance}
\begin{equation}\label{eq:lossinstance}
    \resizebox{0.9\hsize}{!}{%
        $L_{ins}(I_k) = -\log \frac{\exp{(\cos(E_V(I_k),E_T(P_{I_k}^c))/\tau_{ins})}}{\sum_{i=1}^B \exp{(\cos{(E_V(I_i),E_T(P_{I_i}^c)}/\tau_{ins})}},$%
        }
\end{equation}
where $\tau_{ins}$ is the temperature parameter.
In $L_{ins}$ the $B$ prompts $\{P_{I_i}^c,i=1, \cdots, B\}$ are generated by $B$ images while in $L_{domain}$ the $K$ prompts $\{P_I^j,j=1,\cdots,K\}$ are generated by a single image, which is different. The difference of $P_{I_1}^c,P_{I_2}^c,\cdots,P_{I_B}^c$ lies in their instance-specific prompts.

Two loss functions are optimized independently. When optimizing $L_{domain}$, the instance prompt learner $h_\theta$ is frozen. When optimizing $L_{ins}$, we do not change the parameters $P_S$ and $P_G$ as well. This will prevent specific parts of the prompt from containing semantic information that the other parts are intended to represent.

\subsection{Visual Alignment}
With the tuned prompts obtained in the first stage, we train a prompt-based visual aligner $g_\theta$ to align the domain-specific information, where $\theta$ is the parameter. $g_\theta$ takes an image $I$ as the input and maps it to a pre-selected domain $u$, which is denoted as 
\begin{equation}
I^\prime=g_\theta(I).
\end{equation}
Although lacking paired images to train the image-to-image map, we can obtain the prompt descriptions $P_I^u$ of the desired output image $I^\prime$. 
\begin{equation}
    P_I^u=[P_F^1][\textcolor[rgb]{1,0,0}{P_C}][\textcolor[rgb]{0,0,1}{P_S^u}][\textcolor[rgb]{0,1,0}{P_G}] [P_F^2].
\end{equation}
$P_I^u$ consists of the domain-specific prompt of domain $u$ and maintain the instance-conditional and global description of $I$. 
Thus, $g_\theta$ can be optimized by matching the visual embedding of $I^\prime$ and prompt description given by $E_V$ and $E_T$.

Aside from matching the global visual semantic information with the prompt $P_I^u$, we also want the local features of the output image to match the text description. Moreover, VGG context is applied to keep semantic information stable through the alignment as well. Thus, the total loss of the visual aligner consists of global match loss, patch match loss and feature loss.

\textbf{Global match loss:} 
Given the input image $I$ and its $K$ prompts generated by the prompt learner, we aim to match $g_\theta(I)$ and $P_I^u$. The global loss $L_{global}(I)$ is formulated in Eq \eqref{eq:alignedglobal}
\begin{equation}
    L_{global}(I) = 1 - \frac{E_T( P_I^u))\cdot E_V(I^\prime)}{\parallel E_T(P_I^u)\parallel \cdot \parallel E_V(I^\prime)\parallel },
    \label{eq:alignedglobal}
\end{equation}
where $g_\theta$ is the visual aligner network with the parameter $\theta$.

\textbf{Patch match loss:}
We utilize the random crop in $I^\prime$ to divide the image into patches and rotate them with a random angle. Then we calculate the cosine-similarity loss in M small patches $I^\prime_i,i=1,2,\cdots,M$ to match the prompt $P_I^u$.\useshortskip

\begin{equation}\label{model3_coef}
    \resizebox{0.91\hsize}{!}{%
        $\begin{aligned}
    \setlength\abovedisplayskip{0pt}
    L_{patch}(I^\prime) &= \sum_{i=1}^M L_{patch}^i(I^\prime)\\
    &= \sum_{i=1}^M 1 - \frac{E_T( P_I^u))\cdot E_V(I_i^\prime)}{\parallel E_T(P_I^u)\parallel \cdot \parallel E_V(I_i^\prime)\parallel }.
    \end{aligned}$%
        }
\end{equation}
M is the number of patches.

\textbf{Threshold rejection:} The semantic information is not evenly distributed in the image. Some patches in the image are difficult to match with the text embedding because of the lack of semantic information. To solve this problem, we refer to the threshold rejection technique\cite{styleclip,clipstyler}, which will discard the patches whose visual embeddings are far from the text embedding. With a threshold $\tau_{patch}$, the patch loss is formulated as 
\begin{equation}
    L_{patch}^i (I^\prime,\tau_{patch}) = \max(0,L_{patch}^i - \tau_{patch}).
    \setlength{\belowdisplayskip}{3pt}
\end{equation}

\textbf{Feature Loss:} 
We leverage the feature extracted by a pretrained VGG to calculate the content loss $L_{feature}$ by calculating the mean-square error between features extracted by the VGG\cite{vgg} pretrained base on ImageNet\cite{imagenet}. The $L_{feature}$ is formulated as  \useshortskip

\begin{equation}
\setlength{\abovedisplayskip}{-1pt}
    L_{feature}(I) = \parallel VGG(I) - VGG(I^\prime)\parallel.
\end{equation}
\useshortskip
\textbf{Total Loss:} Total Loss is the weighted average of $\textbf{Global match loss},\textbf{Patch match loss},\ and \ \textbf{Feature Loss}$
\begin{equation}
\setlength{\abovedisplayskip}{1pt}
    L_{total} = L_{global} + \lambda_{patch} L_{patch} + \lambda_{feature}L_{feature}.
\setlength{\belowdisplayskip}{1pt}
\end{equation}
\subsection{Robust Policy Training}
To verify our methods, we design a vision-based control task, which is trained by Promxial Policy Optimization(PPO\cite{ppo}). 

To reduce the number of interactions with the environment, we pretrain a feature extractor to obtain a compressed representation from the image by adapting the idea of auto-encoder\cite{autoencoder}. We pretrain a feature extractor based on $D_{policy}$ by sending the image to visual aligner $g_\theta$. A policy head is trained with PPO with the pretrained feature.

\begin{figure}[t]
\centering
\subfigure[ClearNoon]{
    \centering
    \includegraphics[width = 0.4 \columnwidth]{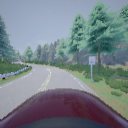}
}
\subfigure[HardRainNoon]{
    \centering
    \includegraphics[width = 0.4 \columnwidth]{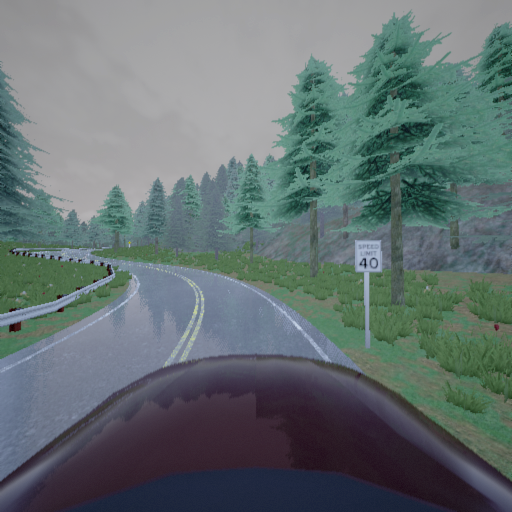}
}
\subfigure[\scalebox{0.9}{WetCloudySunset}]{
    \centering
    \includegraphics[width = 0.4\columnwidth]{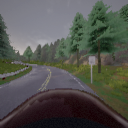}
}
\subfigure[SoftRainSunset]{
\centering
    \includegraphics[width = 0.4\columnwidth]{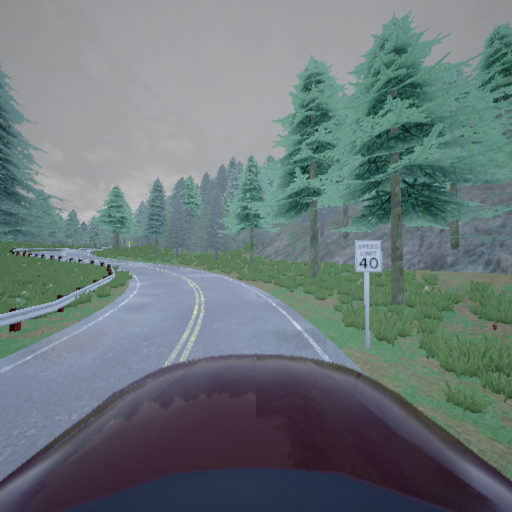}
}
\caption{Illustration of different domains. ClearNoon and HardRainNoon are used to tune the prompt and train the visual aligner. We validate the agent's performance on WetCloudySunset, ClearSunset, and SoftRainSunset, which do not appear in the training stage.}
\label{fig:diffweather}
\end{figure}

\section{Experiment}

\subsection{Setup}
In this section, we will deliver the details of the experiments to validate the agent's generalization ability in CARLA\cite{CARLA}. 
\haihan{CARLA is an autonomous driving simulation engine that provides several maps including urban and suburban areas under different weather conditions and is used to test the adaptation performance of the autonomous driving agent. CARLA also provides generic sensors such as the RGB camera and Radar to generate multi-modal sensational input like the point cloud and BEV image. We use the RGB image captured by a camera fixed on the top of the vehicle to train an autonomous driving agent and validate its performance under different weather conditions}
Various weather conditions are illustrated in Figure \ref{fig:diffweather}.

We build a reinforcement environment based on the CARLA simulator.
The observation space consists of two parts: an image captured by the RGB camera fixed on the car and the current velocity of the vehicle. Action space is a three-dimensional vector, two dimensions control the direction of the car and the other dimension corresponds with the car's acceleration. We have to control the vehicle from the start point to the destination. The environment will terminate when encountering one of the following termination conditions 
     1) The car has arrived at the destination.
     2) The car collapses with other objects. 
     3) The driving time exceeds the time limit.
     4) The car crosses the lane line.
A negative punishment will be given if events 2), 3), and 4) occur and a positive reward will be given if the car arrives at the destination before the time limit. We also want the vehicle to arrive at the destination as quickly as possible. For this reason, a constant negative reward will be given for every step. 
Each domain represents a specific type of weather, which is controlled by a set of parameters, including cloudiness, precipitation, sun altitude angle, and sun azimuth angle.

\subsection{Design of the model and details of training}
\textbf{Design of the Instance Prompt Learner $h_\phi$}: Instance Prompt Learner $h_\theta$ generates a sequence of learnable tokens directly from the image. $h_\theta$ consists of a ResNet19\cite{resnet} and $L_C$ MLP heads. The heads take the feature extracted by ResNet as the input and convert it into $L_C$ instance tokens. 

\textbf{Design of the Visual Aliger  $f_\theta$}: We apply UNet\cite{unet} as the visual aligner, UNet is composed of an encoder and a decoder and has a symmetric structure. The encoder extracts features from the raw image by a CNN layer with $3\times 3$ kernel. Pooling and Skip-connection are also applied in the encoder to reduce the dimension of features and fuse different layers. The decoder will recover the resolution from the feature map via upsampling. We constrain the UNet by aligning the visual representation of the transferred image with the tuned text presentation corresponding to a specific domain.

\textbf{Details of training:} With the help of the pretrained CLIP, A small dataset collected in various domains is applied in prompt tuning and visual alignment. In the experiment, we sample only 100 images from 2 different domains to extract semantic information and build the visual aligner. \haihan{The length of global, domain-specific, and instance prompts are 10, 5, and 10. We also applied different temperatures and learning rates to align the learnable prompts with the visual inputs. For general and domain-specific prompts, the temperature and learning rate are 0.5 and 0.0004. For instance-specific prompts, the temperature and learning rate are 0.1 and 0.00005. The prompt tuning took 4 hours and visual alignment took about 8 hours, both were conducted in A8000. }


\subsection{Experiment Result} 
\TODO{In this part, we evaluate our method from two perspectives. Firstly, we compare it with other state-of-the-art representation-based methods that utilize the experimental settings described in the original article. Moreover, we reduce the training data of these methods to assess the generalization capability of our approach.}
\subsubsection{Comparison with existing methods}
We compared our method with representation-based methods, such as LUSR\cite{LUSR}, CURL\cite{CURL}, and DARLA\cite{darla}. These methods require data from various domains to pretrain a feature extractor and obtain a latent 
domain-shared representation of the visual inputs. The agent derives corresponding actions from the pretrained latent features. These methods aim to extract the domain-agnostic representation across different domains which can generalize to unseen domains. There are also several approaches that train an image-to-image translator, such as CycleGAN-RL\cite{cyclegan}, AdaIN\cite{instancestyletrans}, pix2pix\cite{pix2pix}, and UNIT4RL\cite{transferimagerl}. 
These approaches transform images from the source domain to the target domain.

The performance is illustrated in Table \ref{tab:performance}. Our method exceeds other approaches on unseen domains. Our method utilizes fewer images from multiple domains than other approaches illustrated in Table \ref{tab:performance}. LUSR separates the latent representation into domain-agnostic and domain-specific parts and could generalize well on seen domains. DARLA utilizes $\beta-$VAE to learn a domain-agostic representation by adapting $\beta$. CURL uses data augmentation and contrastive learning to learn the adaptable representation. We apply random crop and grayscale to create positive pairs. From the experiment results, we observe there exists performance degradation in the unseen domain. For image-translation approaches, we find it could only handle the adaptation between two domains. Our approaches achieve higher generalization performance by alleviating the domain bias with less requirement for cross-domain images.
In the baseline setting, we sampled 10000 images per domain to learn a cross-domain invariant representation or the image translator.

\begin{table*}[h]
\centering
\begin{tabular}{cccccc}
    \toprule
     \multirow{2}{*}{method} &\multicolumn{2}{c}{seen domains} & \multicolumn{3}{c}{unseen domains} \\
          \cmidrule(lr){2-3} \cmidrule(lr){4-6}
       & ClearNoon & HardRainNoon & ClearSunset & WetCloudySunset & SoftRainSunset \\ 
    \midrule
        LUSR & 1997.4 & 775.6 & 2079.0 & 582.8 & 533.8 \\ 
        DARLA & 1207.2 & 1257.4 & 450.3 & 162.5 & 1060.3 \\ 
        CURL & 1508.1 & 1308.4 & 801. & 998.5& 	1046.  \\ 
        UNIT4RL & 1636.4 & 322.7 & 298.1 & 239.8 & 383.5 \\
        CycleGAN & 1981.9 & 1528.8 & 277.0 & 637.4 & 657.1 \\ 
        pix2pix & 1821.7 & 520.5 & 157.5 & 420.9 & 662.9 \\ 
        AdaIN & 1829.6 & 285.8 & 499.9 & 101.7 & 245.3 \\ \midrule
        PVA(ours) & \textbf{2004.6} &\textbf{1825.9} & \textbf{2178.4} & \textbf{1775.65} & \textbf{1789.2} \\ 
        \bottomrule
\end{tabular}
\vspace{-0.5em}
    \caption{Domain Generalization Performance of PVA and other benchmarks in CARLA simulation task. ClearNoon and HardRainNoon is included in seen domains. The policy is trained in ClearNoon.}
\vspace{-1em}
    \label{tab:performance}
\end{table*}


We illustrate the inputs of RL agent, which is a latent representation compressed from the raw image. Data augmentation and representation-based methods suffer from distribution shifts between seen and unseen domains. Our approach mitigates the domain bias and maps all domains into a unified distribution.


\begin{figure}[t]
\centering
\includegraphics[width=\columnwidth]{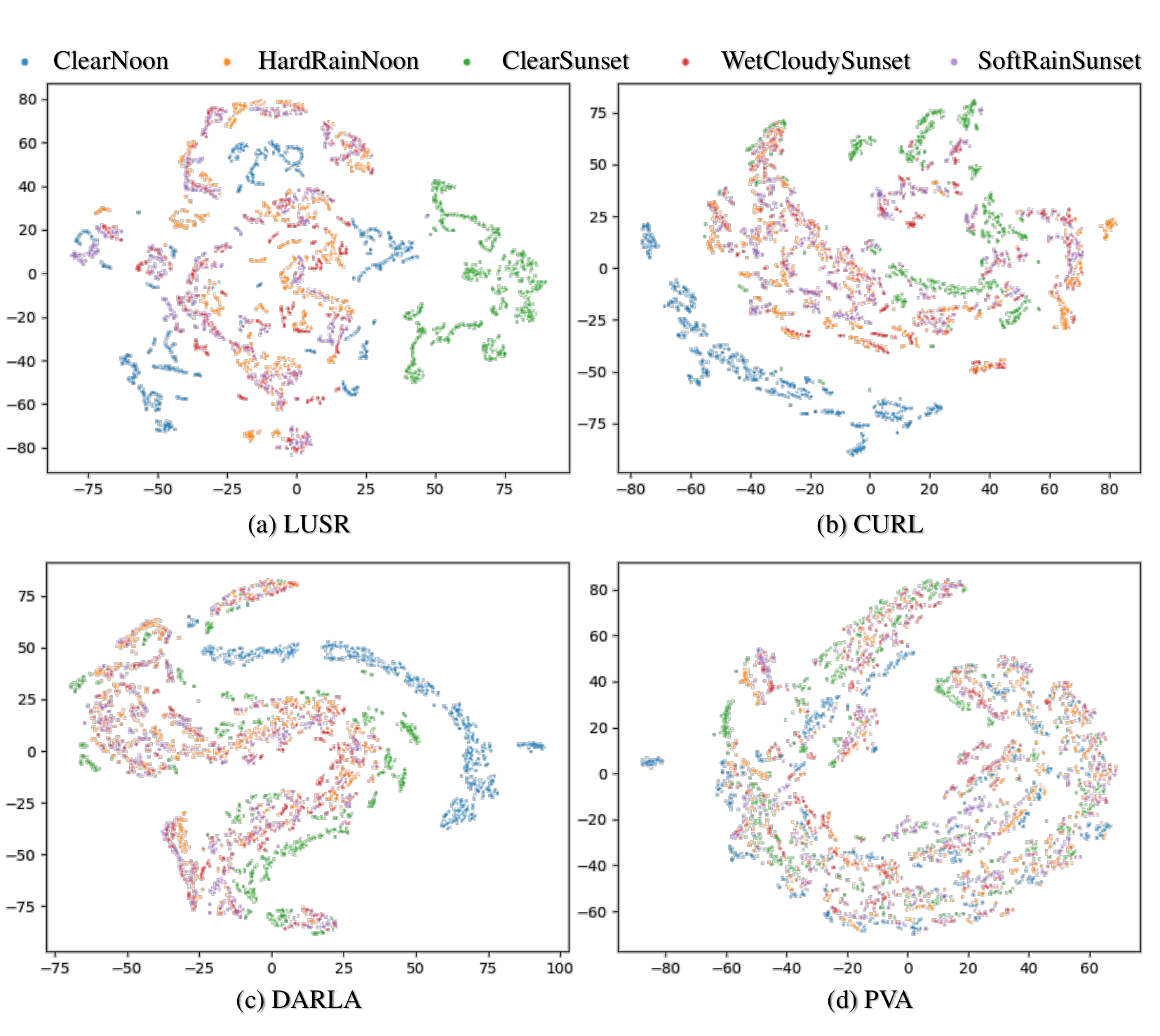}
\caption{Visualization distributions of different domains' latent embeddings, which serve as the input of the RL agent. Different colors are used to represent various domains respectively. We observe our approach alleviates the domain bias between domains, which makes the agent generalize under detrimental observation differences.}
\vspace{-1em}
\label{fig:latentdis}
\end{figure}

\begin{table}[H]
    \centering
   \begin{tabular}{ccc}
    \toprule
      method & ClearNoon &  ClearSunset  \\ 
    \midrule
        CURL & 1227.4 & 199.5  \\ 
        AdaIN & 1420.8 & 313.9 \\ \midrule
        PVA(ours) & \textbf{2004.6} &\textbf{1825.9} \\ 
        \bottomrule
\end{tabular}
\vspace{-0.5em}
    \caption{Compared with other approaches with the same quantity of images.}
    \vspace{-2em}
    \label{tab:same_amount}
\end{table}


\subsubsection{Comparision with other baselines with the same number of images in training}
\TODO{This experiment is conducted in order to verify baseline performances under limited access to various domains.}
Table \ref{tab:same_amount} shows the performance of baselines and our approach under the same amount of images used in the training stage. 
\TODO{We applied 100 images from ClearNoon and HardRainNoon for all baselines illustrated in Table \ref{tab:same_amount}. For CURL, we apply contrastive learning to learn a representation from these images.}
In AdaIN, the small dataset is used to train an image translator between two domains. 
\TODO{our approach outperforms both two baselines under this situation, which indicates the representation-baseline approaches fail to learn the generalizable latent features and the image-to-image method could not generalize to the third domain under limited access to cross-domain data. }

\subsection{Ablation studies}


\textbf{Ablation study of different components of the prompt:} In this part, we conduct experiments to show the necessity of each component in our methods,.
The results are illustrated in Table \ref{tab:ablationstudy}. We find that both three components of prompts are beneficial for the adaptation performance. 
\TODO{Table \ref{tab:ablationstudy} illustrates the influence of different prompts, where the last line corresponds with the default settings. In the table, we apply green, blue, and red tokens to represent global, domain-specific, and instance-level prompts. The first line illustrates the performance without prompt tuning, where a fixed sentence is directly provided in the second stage. In the middle three lines, we remove each part separately and find that the domain-specific prompt has the most influential impact on the final results. }

\textbf{Ablation study of loss function in $L_{total}$}. \haihan{In this part, we discuss the influence of different components in $L_{total}$ when training the visual aligner $f_\theta$. The result is illustrated in Table \ref{tab:ablationloss} .} 
\TODO{In the last three lines, we find that removal of the feature loss will influence the generalization performance dramatically and removals of our two loss functions will bring detrimental impacts on unseened domains. }






\begin{table*}[ht]
\centering
\scalebox{0.95}{
\begin{tabular}{cccccc}
    \toprule
     \multirow{2}{*}{prompt} &\multicolumn{2}{c}{seen domains} & \multicolumn{3}{c}{unseen domains} \\
          \cmidrule(lr){2-3} \cmidrule(lr){4-6}
       & ClearNoon & HardRainNoon & ClearSunset & WetCloudySunset & SoftRainSunset \\ 
    \midrule
        fixed sentence& 1559.4 & 118.63 & 102.6 & 1141.6 & 856.72 \\
        $\textcolor[rgb]{0,0,1}{[v_{s}^1]\cdots [v_{s}^{L_S}]}$$\textcolor[rgb]{1,0,0}{[v_C^1]\cdots [v_C^{L_C}]} $ & 1157.1 & 1120.7 & 716.0 & 1049.2 & 592.7  \\
        $\textcolor[rgb]{0,1,0}{[v_g^1]\cdots [v_g^{L_G}]}$$\textcolor[rgb]{1,0,0}{[v_C^1]\cdots [v_C^{L_C}]} $ & 1361.3 & 1425.48 & 852.6 & 1330.1 & 1055.2 \\
        $\textcolor[rgb]{0,1,0}{[v_g^1]\cdots [v_g^{L_G}]}$$\textcolor[rgb]{0,0,1}{[v_{s}^1]\cdots [v_{s}^{L_S}]}$ & 952.6 & 110.62 & 305.6 & 613.0 & 283.0 \\
        $\textcolor[rgb]{0,1,0}{[v_g^1]\cdots [v_g^{L_G}]}$$\textcolor[rgb]{0,0,1}{[v_{s}^1]\cdots [v_{s}^{L_S}]}$ $\textcolor[rgb]{1,0,0}{[v_C^1]\cdots [v_C^{L_C}]} $ & 2004.6 &{1825.9} & 2178.4 & {1775.7} & {1789.2} \\ 
        \bottomrule
\end{tabular}
}
\vspace{-0.5em}
    \caption{Ablation study of prompt structure. We observe the prompt tuning in stage 1 is effective in improving the generalization ability of the RL agent by replacing the tuned prompts with a fixed template.}
    \vspace{-1em}
    \label{tab:ablationstudy}
\vspace{1em}
\begin{tabular}{cccccccc}
    \toprule
     \multirow{2}{*}{method} & \multirow{2}{*}{$L_{global}$} & \multirow{2}{*}{$L_{patch}$} & \multirow{2}{*}{$L_{feature}$}  &\multicolumn{2}{c}{seen domains} & \multicolumn{2}{c}{unseen domains} \\
          \cmidrule(lr){5-6} \cmidrule(lr){7-8}
        & ~ & ~ & ~ & ClearNoon & HardRainNoon & ClearSunset & WetCloudySunset  \\ 
    \midrule
        with all loss function & $\checkmark$ & $\checkmark$ & $\checkmark$ & 2004.6 & 1825.9 & 2178.4 & 1775.6 \\ 
        without feature loss & $\checkmark$ & $\checkmark$ & $\times$ & 1175.0 & 147.7 & 111.4 & 83.9  \\ 
        without patch loss & $\checkmark$ & $\times$ & $\checkmark$ & 1420.8 & 1455.8 & 313.9 & 993.8  \\ 
        without global loss & $\times$ & $\checkmark$ & $\checkmark$ & 1533.4 & 1480.5 & 993.8 &  708.1 \\
        \bottomrule
\end{tabular}
\vspace{-0.5em}
    \caption{Ablation study of $L_{total}$. We observe $L_{feature}$ is crucial for adapting in HardRainNoon because it provide enough details during the transfer. $L_{global}$ and $L_{patch}$ are important for zero-shot adaptation on unseen domains.}
    \vspace{-1.5em}
    \label{tab:ablationloss}
\end{table*}

\section{Conclusion}
We propose a visual alignment framework to enhance the generalization ability of reinforcement learning agents by training a visual aligner, which aligns the multi-domain images. With the help of explicit semantic constraints, our approach achieves an excellent zero-shot domain generalization performance.
Our method utilizes a sequence of learnable tokens to represent the semantic information in observations and tunes them with a pretrained VLM by aligning the visual and textual representations. Experiments conducted on a visual autonomous driving task show our approach has a better generalization performance over unseen domains with less access to cross-domain data.
\section*{Impact Statement}
This paper presents work whose goal is to advance the field of Machine Learning. There are many potential societal consequences of our work, none which we feel must be specifically highlighted here.
\section*{Acknowledgement}
This work is partially supported by the National Key R\&D Program of China(under Grant 2023YFB4502200), the NSF of China(under Grants U22A2028, U20A20227, 61925208, 62102399, 62222214, 62341411, 62302483, 62102398, 62372436, 62302478, 62302482, 62302480), Strategic Priority Research Program of the Chinese Academy of Sciences, (Grant No. XDB0660300, XDB0660301, XDB0660302), CAS Project for Young Scientists in Basic Research(YSBR-029), Youth Innovation Promotion Association CAS and Xplore Prize.



\bibliography{example_paper}
\bibliographystyle{icml2024}
\newpage
\appendix
\onecolumn
\clearpage
\setcounter{page}{1}

\onecolumn
\section{Additional Experiment result}
We visualize the visual alignment results and compare it with our image translation approaches, such as CycleGAN\cite{cyclegan} and adain\cite{instancestyletrans}. CycleGAN is widely used to translate between two domains, which utilizes a generator to transfer the source domain image to the target domain. AdaIn utilizes instance normalization to transfer styles of images.

\begin{figure*}[h]
\centering
\subfigure[original input images in different illumination conditions]{
\centering
\includegraphics[width=0.23\columnwidth]{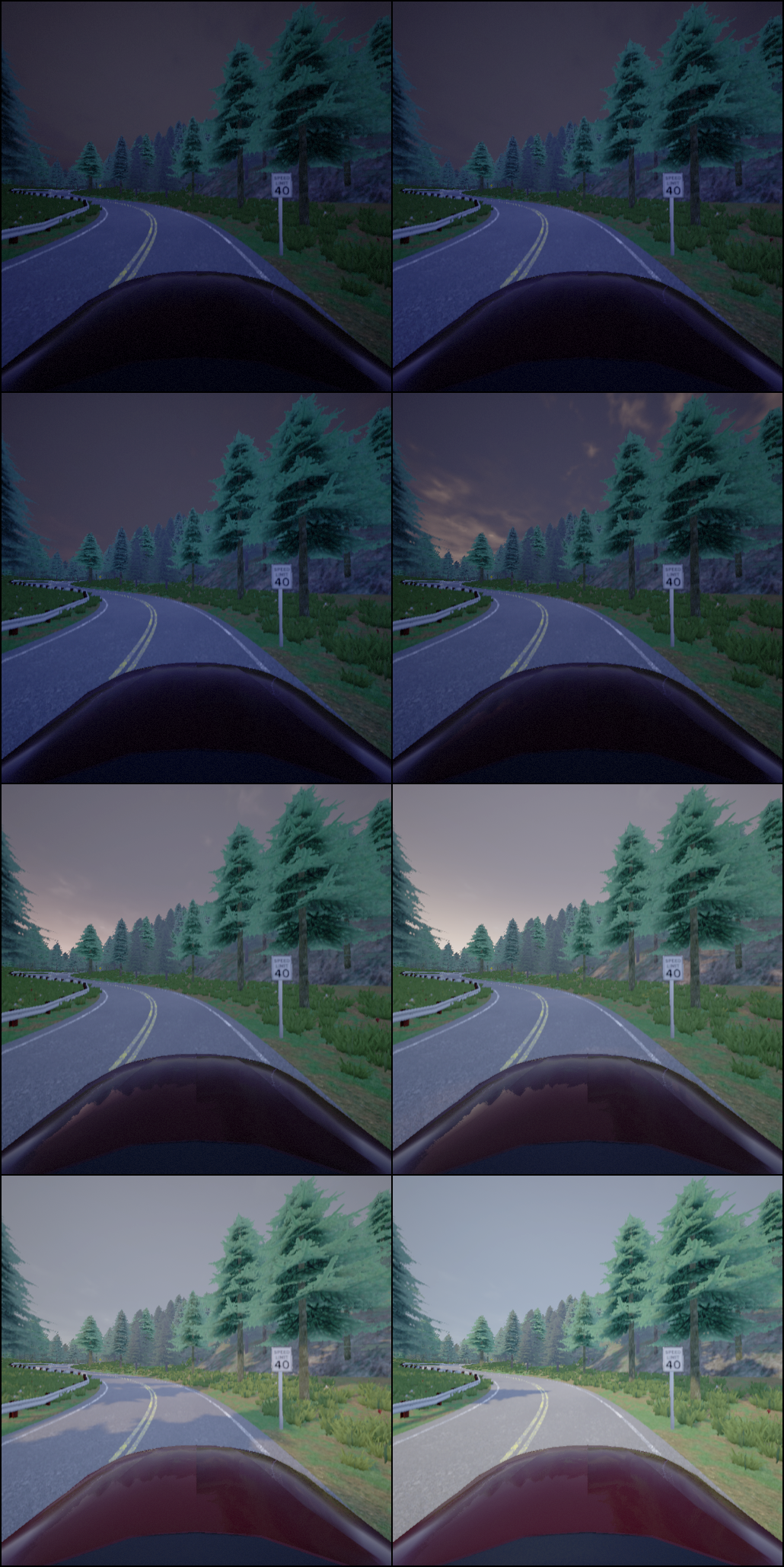}
}
\subfigure[image trans results of the adain\cite{instancestyletrans}]{
\centering
\includegraphics[width=0.23\columnwidth]{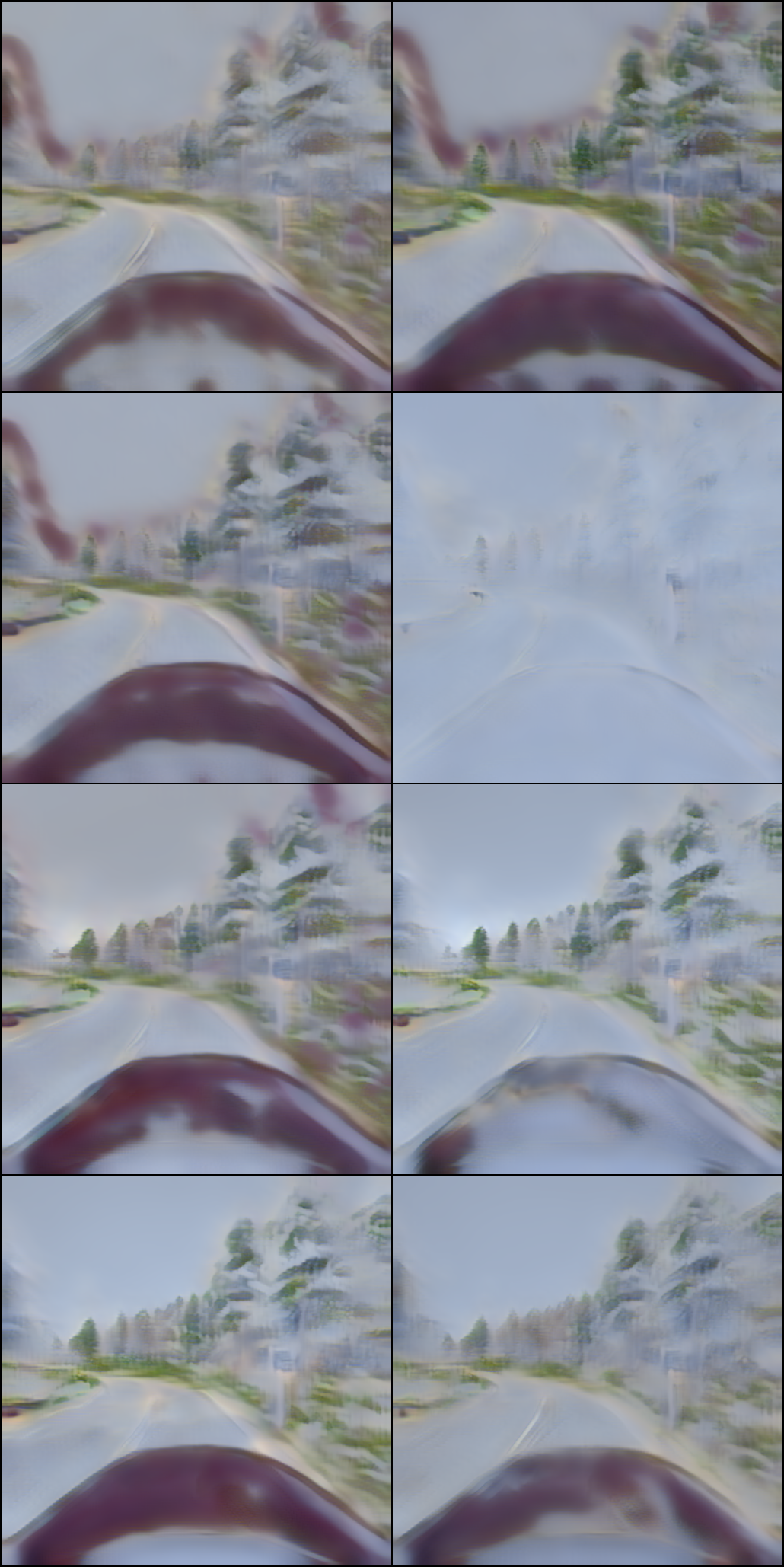}
}
\subfigure[image trans results of the unit\cite{transferimagerl}]{
\centering
\includegraphics[width=0.23\columnwidth]{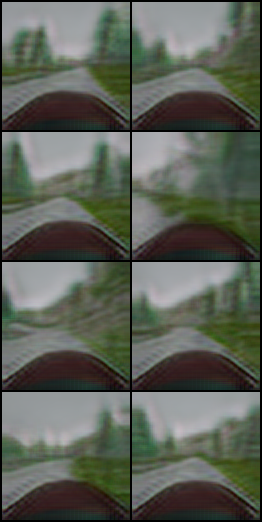}
}
\subfigure[image transfer results of our approach]{
\centering
\includegraphics[width=0.23\columnwidth]{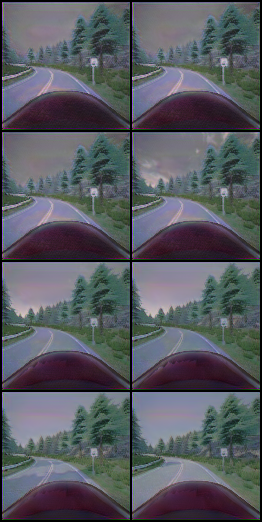}
}
\caption{Compare the visual transfer results of our methods with other image-to-image approaches. The images generated by our method is illustrated in the leftmost sub-figure. You can identify the road line from the transferred images, which is essential for the control task.}
\vspace{-1em}
\label{fig:visualaligner}
\end{figure*}
From the visualization result illustrated in Figure \ref{fig:visualaligner}, we observe our method keeps the details within the image unchanged and decreases the difference in the original input images. 
The images generated by AdaIN and CycleGAN contain blurring and confusion.
\section{Effects of prompt length}
The learnable parts of the prompt consist of global, domain-specific, and instance parts.
We conduct experiments to verify the influence of each part length. 
The default setting is $L_G=10,L_S = 5,L_C = 10$, which represents the length of global prompts, domain-specific prompts, and instance-conditional prompts. 
We illustrate the experiment result in the following tables. In Table \ref{tab:LCinfluence}, Table \ref{tab:LGinfluence}, and Table \ref{tab:LSinfluence}, we show the influence of instance-conditional, global, and domain-specific prompts length. 

For the instance-conditional prompt $P_C$, experiments indicate that too short instance-conditional prompts will make the prompts insufficient to describe the semantic information in the image and will cause the RL algorithm to fail in transferred images. 

For the global prompt shared by all domains $P_G$,  we also observe similar phenomena. However, this does not mean the longer the prompt, the better the performance of the agent. From Table \ref{tab:LSinfluence}, we observe that when $L_S$ exceeds 5, the performance actually degrades. The possible reason might result from the difficulty of optimizing overly long prompts.

\begin{figure*}[b]
  \centering  
    
  \subfigure[The length of instance prompts is 1.]{
    \centering  
    \includegraphics[width=0.6\columnwidth]{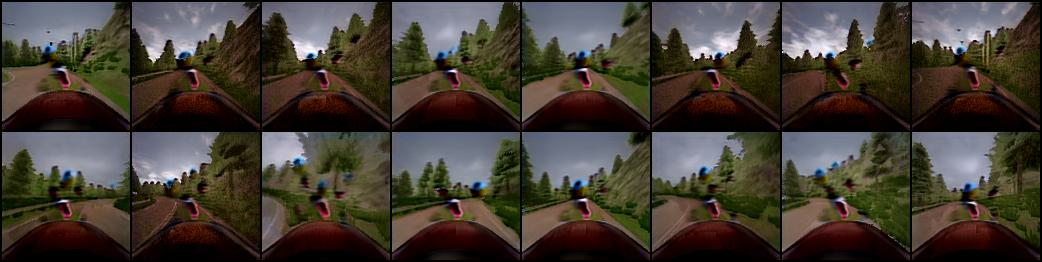}  
  }
    
  \subfigure[The length of instance prompt is 7.]{
    \centering  
    \includegraphics[width=0.6\textwidth]{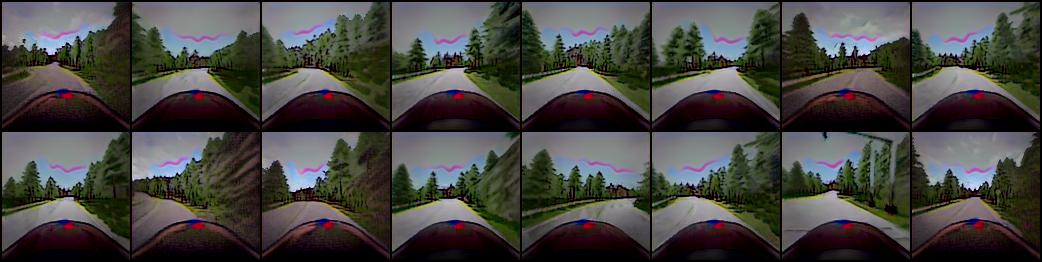}  
  }
    
  \caption{Visual transfer result of different instance prompt length}  
  \label{fig:transferdifferentinstance}  
  \centering  
    
  \subfigure[The length of domain-specific prompts is 1.]{
    \centering  
    \includegraphics[width=0.6\columnwidth]{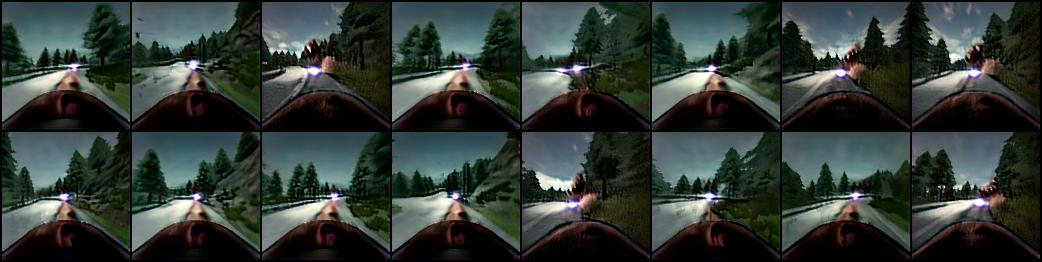}  
  }
    
  \subfigure[The length of domain-specific prompt is 7.]{
    \centering  
    \includegraphics[width=0.6\textwidth]{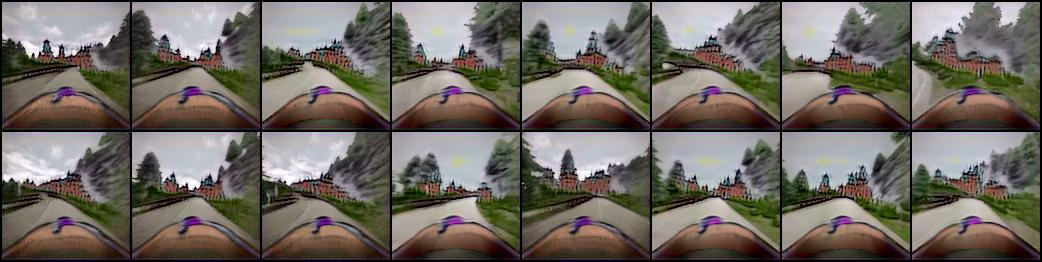}  
  }
    
  \caption{Visual transfer result of different domain-specific prompt length}  
  \label{fig:transferdifferentdomain}  
  \centering  
    
  \subfigure[The length of global prompts is 1.]{
    \centering  
    \includegraphics[width=0.6\columnwidth]{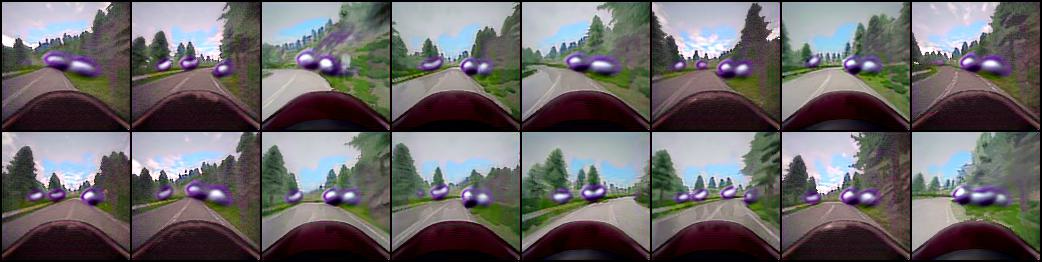}  
  }
    
  \subfigure[The length of global prompt is 10.]{
    \centering  
    \includegraphics[width=0.6\textwidth]{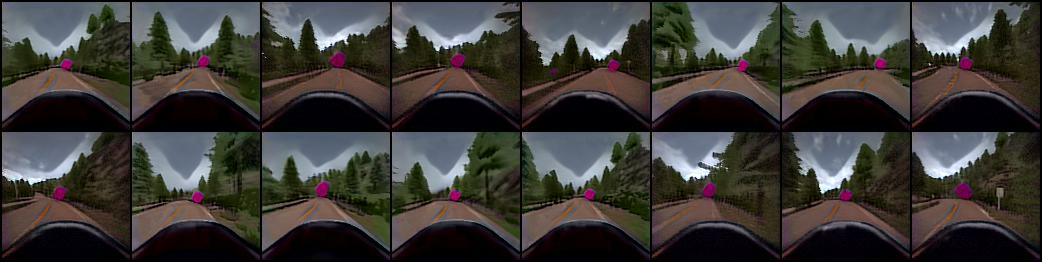}  
  }
    
  \caption{Visual transfer result of different global prompt length}  
  \label{fig:transferdifferentglobal}  
\end{figure*}

We illustrate the image transfer results under different settings in prompt lengths in Figure  \ref{fig:transferdifferentinstance}, Figure \ref{fig:transferdifferentdomain}, and Figure \ref{fig:transferdifferentglobal}. From these figures, we find improving the prompt length will mitigate distortion in aligned images.

\section{Details of the environment}
In this section, we will discuss the details of the environment, including its observation, action, and reward function. The observation is a 512$\times$512 size image captured by an RGB camera fixed on the top of the car facing forward. The action space is a two-dimension vector $(acc,steer)$, where $acc\in(-3,3)$ controls the acceleration of the vehicle moving forward or backward and $steer\in(-0.3,0.3)$ controls the amplitude of the vehicle's left and right turning.

The reward function is the weighted average of the following items.
\begin{itemize}
    \item The speed of the vehicle $v$. We want to arrive at the destination as fast as possible
    \item A negative reward $col$ will be given when a collision event happens.
    \item A negative reward $out$ will be given when the vehicle moves outside the lane.
    \item A negative reward $r_{const} = -0.1$ will be given every step
\end{itemize}
The reward function R is formulated as
\begin{equation}
    R = \lambda_v \times v+\lambda_{col} \times col + \lambda_{out}\times out + r_{const}
\end{equation}
In the experiments, we select $\lambda_v =1, \lambda_{col} = \lambda_{out}=100$
\section{Validation of our approach in other reinforcement learning tasks}
We validate the effectiveness of our framework in other visual-based reinforcement learning task, such as mujoco. We alter the background to change the domain context.
\begin{figure*}[h]
\centering
\subfigure[bear]{
\centering
\includegraphics[width=0.23\columnwidth]{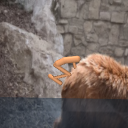}
}
\subfigure[bike]{
\centering
\includegraphics[width=0.23\columnwidth]{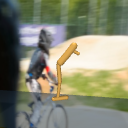}
}
\subfigure[bmx]{
\centering
\includegraphics[width=0.23\columnwidth]{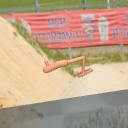}
}
\subfigure[boat]{
\centering
\includegraphics[width=0.23\columnwidth]{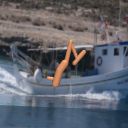}
}
\caption{Different Background of the mujoco task.}
\vspace{-1em}
\label{fig:mujoco_bg}
\end{figure*}
We validate our approach in three mujoco tasks(cartpole-balance/cheetah-run/point-mass) and compare it with CURL. The experiment result indicates our framework outperforms other baseline.
\begin{table*}[h]
\centering
\begin{tabular}{cccccc}
    \toprule
     \multirow{2}{*}{Method} &\multirow{2}{*}{task name} & \multicolumn{4}{c}{domain name} \\
           \cmidrule(lr){3-6}
       &  & HardRainNoon & ClearSunset & WetCloudySunset & SoftRainSunset \\ 
    \midrule
        PVA & cartpole-balance & 967.8 & 914.2 & 919.7	& 968.8 \\
        CURL & cartpole-balance & 529.8	& 424.5	& 429.9	& 386.6 \\
        PVA & cheetah-run & 212.9	& 207.7	& 214.0	& 216.5 \\
        CURL & cheetah-run &178.2	& 166.7	& 161.2	& 164.1 \\
        PVA & point-mass & 855.2 &	849.1 &	848.7 &	842.9 \\
        CURL & point-mass &279.3 &	157.3 &	186.2 &	169.5 \\
        \bottomrule
\end{tabular}
\caption{Performance of our approach in mujoco tasks.}
\label{tab:baselines}
\end{table*}
\section{Additional experiment result to prove that different parts within the prompt are related to various }
In this part, we conduct experiments to demonstrate the correlation between different parts of prompts and visual semantic information. In detail, in order to show the relevance of a given prompt and the corresponding image in the pixel level, we calculate the CLIP score between the prompt and whole image. Then we back propagate the clip score to the original image and obtain gradient with the same size of the input image, which reflect the contribution of each pixel for the semantic matching. In other words, a small disturbance in the patches with high gradient will impact the semantic matching score dramatically. The visualization results is illustrated in following images.

\begin{figure*}[h]
\centering
\subfigure[original image of different domains(HardRainNoon, ClearNight and ClearNoon)]{
\centering
\includegraphics[width=0.3\columnwidth]{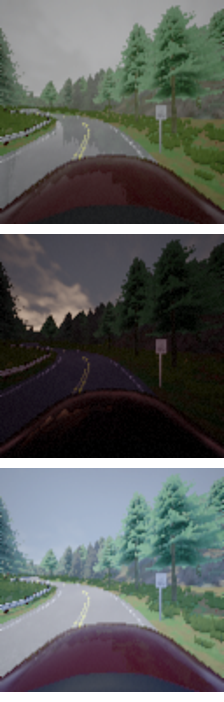}
}
\subfigure[gradcam gradient figure between domain-agnostic prompt and original image]{
\centering
\includegraphics[width=0.3\columnwidth]{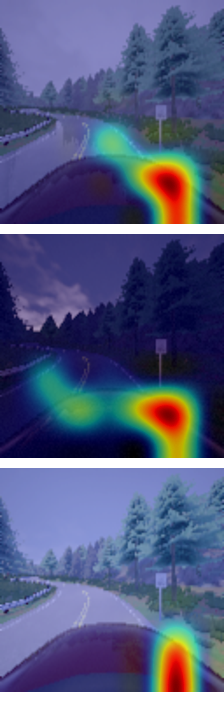}
\label{fig:gradcam_agnostic}
}
\subfigure[gradcam gradient figure between domain-aware prompt and original image]{
\centering
\includegraphics[width=0.3\columnwidth]{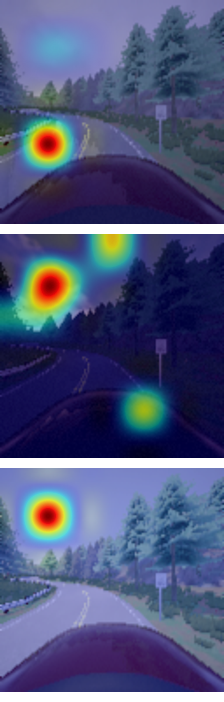}
\label{fig:gradcam_domain_aware}
}
\caption{GradCAM visualization result.}
\vspace{-1em}
\label{fig:gradcam}
\end{figure*}
In the Figure \ref{fig:gradcam}, we visualize the gradcam gradient map between domain-agnostic/domain-specific prompts and input images(as illustrated in \ref{fig:gradcam_agnostic} and \ref{fig:gradcam_domain_aware}). Here is an intriguing finding from GradCAM images. According to Figure \ref{fig:gradcam_domain_aware}, the hot-spot of GradCAM images locates at sky in domain ClearNoon and ClearNight. However, the case is different in domain HardRainNoon, where the hot-spot locates at road. By comparing the original image, we find a reflection caused by standing water on the road in the image, which contains domain-related information(weather). As for the domain-agnostic information, the hot-spot area locates at the car, which indicates this part is related to the position and pose of the vehicle. 

\begin{table*}[h]
\centering
\begin{tabular}{cccccc}
    \toprule
     \multirow{2}{*}{$L_C$} &\multicolumn{2}{c}{seen domains} & \multicolumn{3}{c}{unseen domains} \\
          \cmidrule(lr){2-3} \cmidrule(lr){4-6}
       & ClearNoon & HardRainNoon & ClearSunset & WetCloudySunset & SoftRainSunset \\ 
    \midrule
        1 & 240.4 & 210.5 & 24.4 & 219.5 & 214.0 \\
        3 & 1829.9 & 2122.6 & 1676.0 & 2104.1 & 1977.6 \\
        7 & 1929.3 & 2180.1 & 1913.1 & 1524.2 & 2065.2 \\
        10 & 2004.6 &{1825.9} & 2178.4 & {1775.65} & {1789.2} \\ 
        \bottomrule
\end{tabular}
\caption{Fix $L_G=10,L_S = 5$ and change the length of instance-conditional prompts.}
\label{tab:LCinfluence}
\end{table*}

\begin{table*}[h]
\centering

\begin{tabular}{cccccc}
    \toprule
     \multirow{2}{*}{$L_G$} &\multicolumn{2}{c}{seen domains} & \multicolumn{3}{c}{unseen domains} \\
          \cmidrule(lr){2-3} \cmidrule(lr){4-6}
       & ClearNoon & HardRainNoon & ClearSunset & WetCloudySunset & SoftRainSunset \\ 
    \midrule
        1 & 1435.4 & 1887.7 & 1677.3 & 1526.5 & 1556.6  \\
        3 & 733.4 & 859.8 & 809.4 & 78.04 & 199.63\\
        7 & 1797.6 & 1805.0 & 560.8 & 1384.84 & 1418.5 \\
        10 & 2004.6 &{1825.9} & 2178.4 & {1775.65} & {1789.2} \\ 
        \bottomrule
\end{tabular}
    \caption{Fix $L_C=10,L_S = 5$ and change the length of global prompts.}
    \label{tab:LGinfluence}
\end{table*}

\begin{table*}[h]
    
\centering
\begin{tabular}{cccccc}
\toprule
     \multirow{2}{*}{$L_S$} &\multicolumn{2}{c}{seen domains} & \multicolumn{3}{c}{unseen domains} \\
          \cmidrule(lr){2-3} \cmidrule(lr){4-6}
       & ClearNoon & HardRainNoon & ClearSunset & WetCloudySunset & SoftRainSunset \\ 
    \midrule
        1 & 1224.37 & 90.6 & 73.9 & 91.5 & 91.4   \\
        3 &  1322.9 & 1206.2 & 826.3 & 1734.0 & 1386.1 \\
        7 &  1420.8 & 1533.4 & 1480.5 & 1468.8 & 1654.9 \\
        5 & 2004.6 &{1825.9} & 2178.4 & {1775.65} & {1789.2} \\ 
        \bottomrule
\end{tabular}
    \caption{Fix $L_C=10,L_G = 10$ and change the length of domain specific prompts.}
    \label{tab:LSinfluence}
\end{table*}
\end{document}